\documentclass[10pt,twocolumn,letterpaper, dvipsnames]{article}

\usepackage{cvpr}              

\usepackage{graphicx}
\usepackage{amsmath}
\usepackage{amssymb}
\usepackage{booktabs}
\usepackage{algorithm}
\usepackage{algorithmic}
\usepackage{adjustbox}
\usepackage{multirow}
\usepackage{xcolor}
\usepackage{placeins}

\newcommand\Tstrut{\rule{0pt}{2.6ex}}         
\newcommand\Bstrut{\rule[-0.9ex]{0pt}{0pt}}   

\usepackage[pagebackref,breaklinks,colorlinks]{hyperref}

\usepackage[capitalize]{cleveref}
\crefname{section}{Sec.}{Secs.}
\Crefname{section}{Section}{Sections}
\Crefname{table}{Table}{Tables}
\crefname{table}{Tab.}{Tabs.}

\def\cvprPaperID{29} 
\def\confName{CVPR}
\def\confYear{2023}

\begin{document}

\title{Continual Domain Adaptation through Pruning-aided Domain-specific Weight Modulation}

\author{Prasanna B\\
{\tt\small bprasanna@iisc.ac.in}
\and
Sunandini Sanyal\\
{\tt\small sunandinis@iisc.ac.in}
\and
R. Venkatesh Babu\\
{\tt\small venky@iisc.ac.in}
\\Vision and AI Lab, Indian Institute of Science, Bengaluru
}
\maketitle



\begin{abstract}
In this paper, we propose to develop a method to address unsupervised domain adaptation (UDA) in a practical setting of continual learning (CL). The goal is to update the model on continually changing domains while preserving domain-specific knowledge to prevent catastrophic forgetting of past-seen domains. To this end, we build a framework for preserving domain-specific features utilizing the inherent model capacity via pruning. We also perform effective inference using a novel batch-norm based metric to predict the final model parameters to be used accurately. Our approach achieves not only state-of-the-art performance but also prevents catastrophic forgetting of past domains significantly. Our code is made publicly available.\footnote{Github Page: \url{https://github.com/PrasannaB29/PACDA}}.   
\end{abstract}

\section{Introduction}
\label{sec:intro}
\noindent
Deep learning models often fail to perform well on unseen data (\textit{target domain}) that is different from their training data (\textit{source domain}) distribution (referred to as \textit{domain-shift}). Domain adaptation techniques \cite{ganin2015unsupervised, tzeng2017adversarial, sun2016deep} seek to address this problem of domain shift by adapting the source model to the new target domain. However, they fail to generalize well in the case of multiple sequentially changing domains. In our work, we explore a challenging setting of continual learning (CL) in domain adaptation, where the goal is to keep adapting the model over several sequential domain shifts. 
For instance, a model trained on clear weather data must adapt to other weather conditions such as snowy, foggy, and rainy to achieve optimal performance \cite{wang2022continual}. Therefore, developing a robust approach toward continual domain adaptation is essential for real-world deployment scenarios. Also, we assume a practical setting of source-free DA \cite{kundu2020universal} where data sharing across domains is restricted due to privacy concerns, as seen in most real-life scenarios.
\begin{figure}
\centering
        \includegraphics[width=1.0\linewidth]{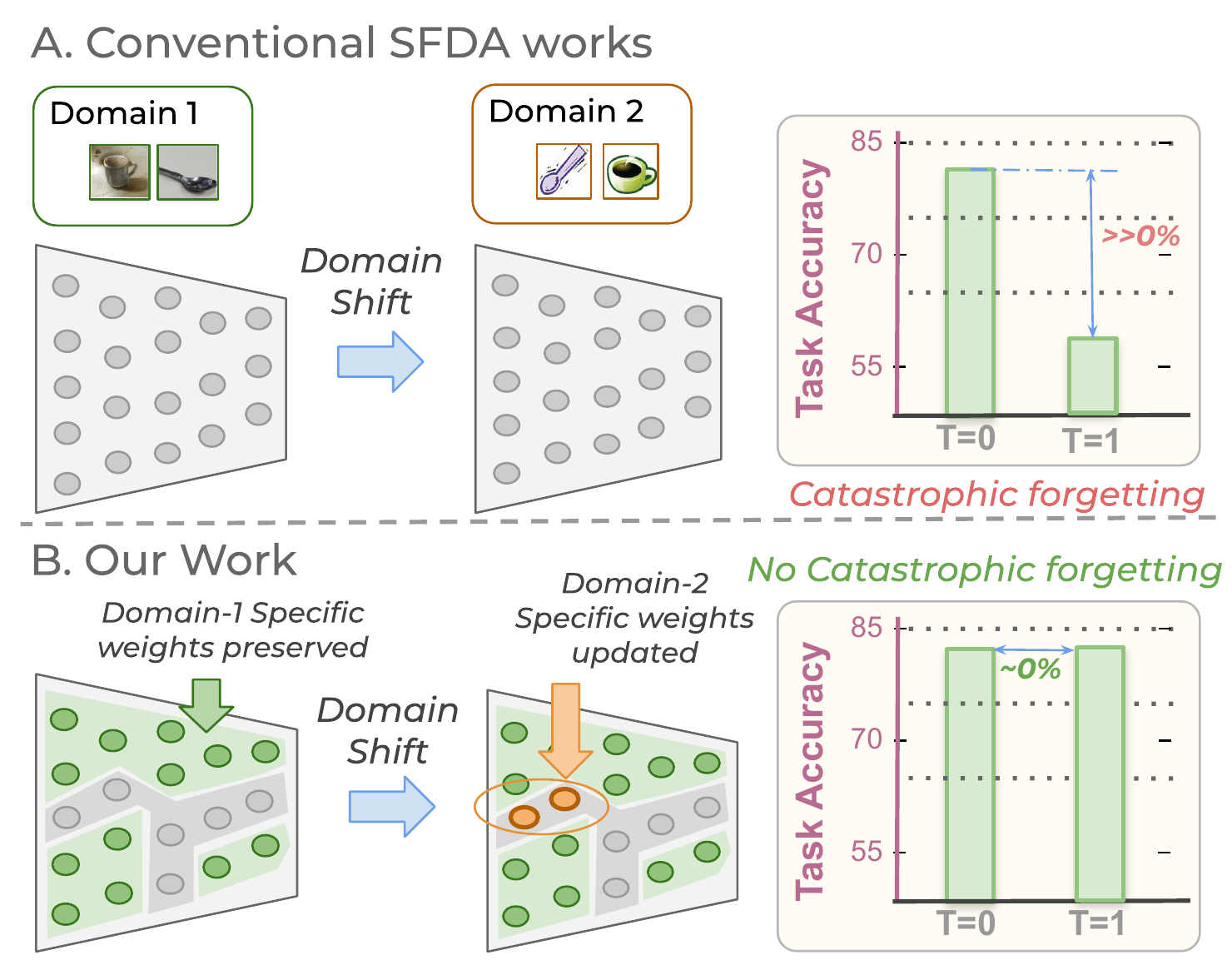}
        \caption{\textbf{A.} Conventional Source-Free Domain Adaptation (SFDA) works do not preserve the source domain performance after domain shift occurs, leading to catastrophic forgetting and a sharp decline in the source domain accuracy. \textbf{B.} Our method preserves domain-specific characteristics inherently within the model to prevent catastrophic forgetting of previously-seen domains.}
        \vspace{-1em}
       \label{fig:teaser}
\end{figure}

\noindent
Prior works on source-free domain adaptation \cite{SHOT++, NRC} aim to enhance the in-domain performance by adapting the source model to the incoming target data. However, such an approach depletes the crucial domain-specific knowledge of the source domain, resulting in catastrophic forgetting (as shown in Fig. \ref{fig:teaser}). As a result, these methods fail to scale 
sequentially changing domains. A possible solution is to store a set of samples from each domain \cite{NEURIPS2021_5caf41d6}, but such methods are ineffective for real-world applications with privacy concerns (where data sharing between parties is restricted) or high memory limitations (e.g., in mobile devices). 

\noindent
In our work, we provide a way to preserve the domain-specific properties within the model and prevent catastrophic forgetting without requiring additional storage. We also learn domain-specific features for a new target domain while maintaining the performance on the previously encountered domains. Since the source domain is inaccessible after source-side training, we explore the inherent potential of the model to preserve domain-specific features using pruning. We find that a fraction of model parameters is sufficient to preserve the domain-specific statistics. Hence, we develop a novel pruning-based algorithm \textbf{P}runing-\textbf{a}ided \textbf{C}ontinual \textbf{D}omain
\textbf{A}daptation (\textbf{PaCDA}) for the challenging task of continual domain adaptation. We also propose a novel \textbf{B}atch \textbf{N}orm \textbf{S}tatistic \textbf{D}eviation (\textbf{BNSD}) metric to evaluate which model parameters to be used during inference for a test domain. 

\noindent
We outline the major contributions of our work as follows:
\begin{itemize}
    \item We investigate and provide pruning-based results on how a model's inherent capacity could be leveraged to store domain-specific features. To this end, we develop our novel framework \textbf{P}runing-\textbf{a}ided \textbf{C}ontinual \textbf{D}omain
\textbf{A}daptation (\textbf{PaCDA}) for enhancing domain-specific knowledge in the model
    \item We define a novel \textbf{B}atch \textbf{N}orm \textbf{S}tatistic \textbf{D}eviation (\textbf{BNSD}) metric for model parameter selection during inference. We also demonstrate the effectiveness of the proposed metric on the Office-Home dataset.
    \item We achieve state-of-the-art performance on a continual DA benchmark, significantly reducing catastrophic forgetting of the past seen domains.
\end{itemize}

\section{Related Works}
\subsection{Source-free UDA}
\noindent
Unsupervised Domain adaptation \cite{xu2021cdtrans, toldo2021unsupervised, zhang2019category, dong2020cscl, mei2020instance,rangwani2022closer} aims to adapt a model trained on a source domain to a new target domain. Source-free domain adaptation (SFDA)\cite{SHOT, NRC} is a more constrained setting where the source domain data is not accessible during target adaptation. 
However, these methods are directly not applicable for a  continual learning setting \cite{NEURIPS2021_5caf41d6} where the goal is to retain performance on previously trained domains.

\subsection{Continual Domain Adaptation}
\noindent
Prior works \cite{NEURIPS2021_5caf41d6, xu2022delving} propose to tackle the novel setting of continual DA where domain shifts occur sequentially. Rostami \ et al. \cite{NEURIPS2021_5caf41d6} use experience replay from a memory buffer that stores a fixed number of confident samples per class per domain. This buffer is appended with the current training data. Our work adheres to the restriction of source-free DA \cite{kundu2020universal} since storing samples for replay is prohibited due to privacy restrictions. On the other hand, GSFDA \cite{GSFDA} does not use experience replay but assumes the availability of domain-id of the inference samples to use appropriate domain attention vectors. Our work PaCDA uses a novel 
strategy to identify the domain of the incoming batch of samples during inference time and does not require prior knowledge of the domain to which inference samples belong. 

\begin{figure}
\centering
  \includegraphics[width=1.0\linewidth]{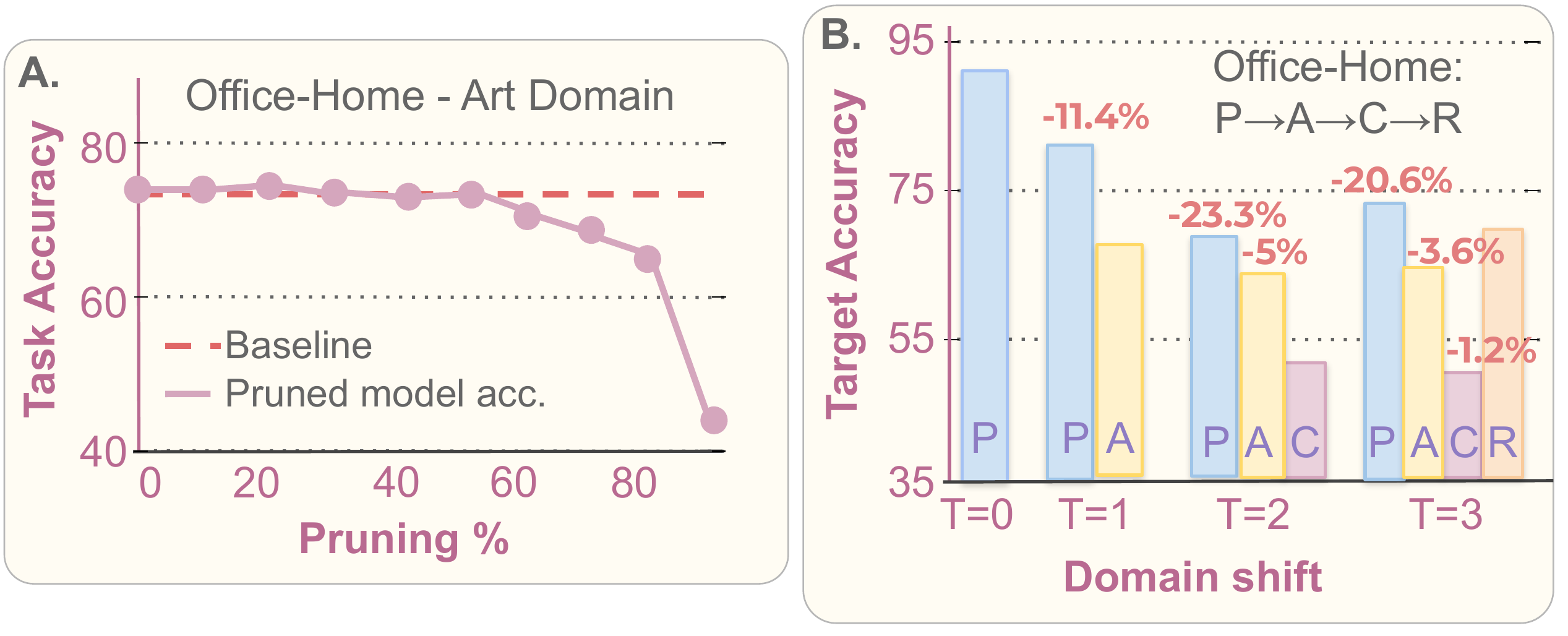}
  \caption{\textbf{A.} Accuracy vs. Pruning \% on a model trained using Art domain data from Office-home. Note that, until $50$\% pruning, the model's accuracy is approximately as good as the original model. The model's performance does not deteriorate significantly until 70\% pruning. 
  \textbf{B.} Catastrophic forgetting of previously trained domains over PACR sequence. Source domain acc. drops by 23.3\%, which significantly impacts performance.}
  \label{fig:pruning_ablation}
  \vspace{-1em}
\end{figure}

\subsection{Pruning for Incremental/Continual Learning}
\noindent
Prior works \cite{li2017learning, lee2017overcoming} use knowledge distillation strategies to minimize forgetting in a task-incremental setting. Pruning-based works \cite{liu2017learning, DBLP:journals/corr/HanPNMTECTD16, DBLP:journals/corr/abs-1711-05769, sui2021chip, zhu2017prune} aim to sparsify and create different network parameters for every new task. 
PackNet \cite{DBLP:journals/corr/abs-1711-05769} uses iterative pruning for learning new tasks and retains the performance on previously learned tasks while training on subsequent tasks. Pruning-based DA works also exist in medical imaging \cite{DBLP:conf/miccai/BayasiHG21} and automatic speech recognition \cite{10022820}. We draw motivation from these prior pruning works and aim to leverage weight-masking for faster domain adaptation without increasing the model's parameters.

\section{Approach}
\subsection{\textbf{Notation}}
\noindent
We consider a labeled source domain dataset 
${{{\mathcal{D}}_s = \{( x_s, y_s) : x_s \!\in\! {\mathcal{X}}, y_s \!\in\! {\mathcal{Y}}\}}}$ where ${\mathcal{X}}$ denotes the input data space and $\mathcal{Y}$ denotes the task label set. We operate in a practical setting of continual DA \cite{NEURIPS2021_5caf41d6} where the source-trained model encounters new target domains sequentially and data-sharing among domains is restricted. Our work follows a practical source-free vendor-client setting \cite{kundu2020universal} where a vendor shares a source model but with multiple clients, one after the other. Hence, the vendor-trained source model encounters different target domains sequentially. Our goal is to update this source-trained model continually on the subsequent target domains  ${{\mathcal{D}}_t = \{ x_t: x_t \!\in \!{\mathcal{X}}\}}$ where $t \in  \{1, 2,...\text{T}\}$, such that the model achieves best performance on all the domains and catastrophic forgetting on past seen domains are minimized. Please note that we assume that all domains share the same task label set $\mathcal{Y}$.


\noindent
Prior works of domain invariance \cite{ganin2016domain} propose to capture domain-invariant features that generalize well to multiple domains. However, an optimal model requires domain-specific knowledge \cite{dubey2021adaptive} to achieve the best performance. Conventional SFDA works \cite{NRC, SHOT, SHOT++} enhance the domain-specific knowledge to improve adaptation performance, but this often leads to catastrophic forgetting of the past seen source domain. As a Baseline, we extend the SHOT approach \cite{SHOT} to a Continual DA setting (shown in Table \ref{table:comparison}), where we adapt to successive domains using SHOT without any mechanism to mitigate catastrophic forgetting. This leads to a sharp decline in the accuracy of past-seen domains, proving the unsuitability of the SFDA methods in a practical scenario of continual DA. In practice, a deployed model often encounters data from different domains sequentially. Hence, it is crucial to leverage the knowledge learned from the old domains to not only improve learning in new domains but also prevent catastrophic forgetting. Therefore, in our work, we propose to address the critical question: \textit{"How do we build a framework for learning continually changing domains while maintaining good performance on all previously-seen domains?"}

\subsection{\textbf{Exploiting the inherent potential of the source model}}
\noindent
In order to build our framework, we propose to preserve the domain-specific knowledge for each domain in the model itself to prevent forgetting. 
The Lottery Ticket Hypothesis \cite{frankle2018lottery} states that a subset of model parameters, i.e., a "winning" ticket, is sufficient for attaining the desired task accuracy. So pruning away the other parameters does not cause a significant performance drop. To investigate this phenomenon in the context of domain adaptation, we first train a model on a source domain (domain Art in Office-Home) and observe the drop in the task accuracy after different percentages of L1-based weight pruning \cite{Pruning_intro_1, DBLP:journals/corr/HanPNMTECTD16}. Figure \ref{fig:pruning_ablation} shows that the task performance declines only after 50\% of the model parameters are pruned. This demonstrates that the inherent redundancy of a neural network could be leveraged to compress the model capacity and utilize the remaining parameters for storing additional domain-specific knowledge encountered in future domains. Hence, this motivates us to explore a pruning-based framework for domain adaptation in a continual DA setting.

\begin{figure*}[t]
\centering
     \includegraphics[width=1.0\linewidth]{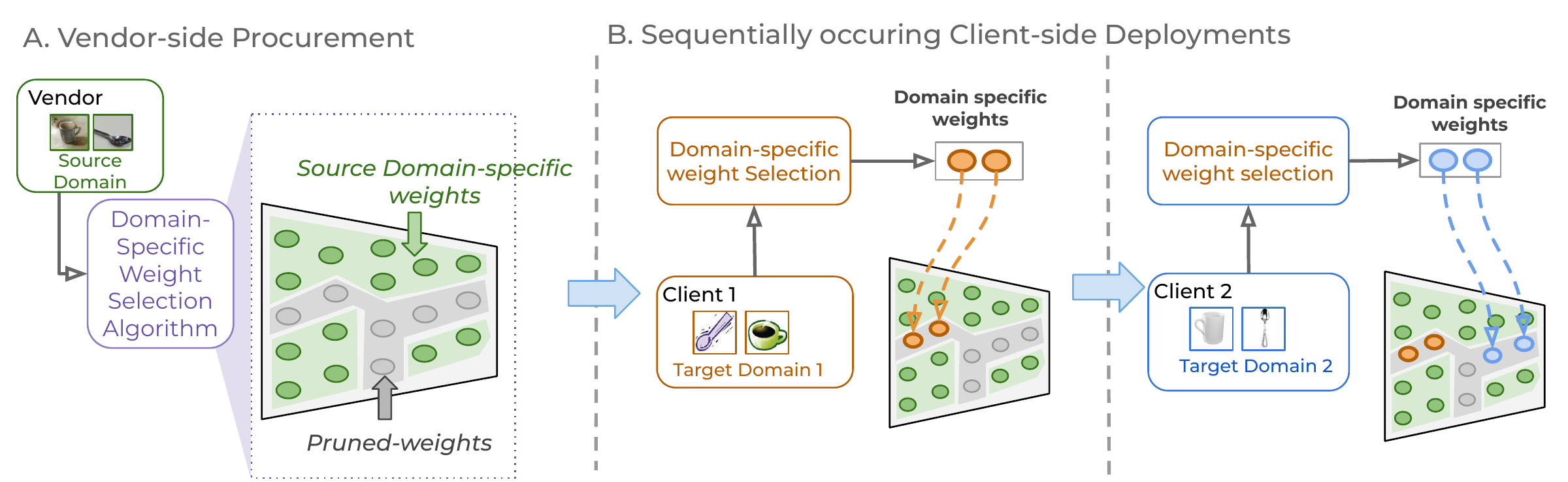}
        \caption{\textbf{PaCDA Approach:} 
        \textit{\textbf{A. Vendor side Procurement: }}The vendor uses the domain-specific weight selection algorithm to select domain-specific weights into the source model. \textit{\textbf{B. Client-side adaptation:}} The vendor-trained Source model is shared with client 1 to update only domain-specific weights from the remaining pruned ones. Client 2 obtains the Client 1 trained model and follows the same procedure. During inference, a batch-norm based metric is used to obtain the final model predictions.
        }
        
       \label{fig:fig_main_approach}
\end{figure*}
\vspace{-0.5em}
        

\subsection{\textbf{Training algorithm}}
\noindent
We propose our novel algorithm \textbf{P}runing-\textbf{a}ided \textbf{C}ontinual \textbf{D}omain \textbf{A}daptation (\textbf{PaCDA}),  
that mainly comprises of Domain-specific weight selection and training phase, and an inference phase. The vendor uses the weight selection algorithm to prepare and share the model with the client. During inference, the client uses a batch norm based inference metric to predict the network mask and identify the model parameters for the incoming data batch. The following sections explain the details of each component of the algorithm. 
\subsubsection{\textbf{Source Model Training}}\label{III.C.1}
The source model consists of two modules: the feature encoding module $g_s: \mathcal{X}\to \mathcal{R}^{d}$ and the classifier module $h_s: \mathcal{R}^{d}\to \mathcal{R}^{K}$, i.e., $f_s(x) = h_s\left(g_s(x)\right)$. Here, $d$ denotes the feature dimension, and $K$ denotes the number of task labels. The source model is trained on labeled source domain data \begin{math}{\mathcal{D}_\mathcal{S}}\end{math} by minimizing the cross-entropy loss as follows:
\begin{equation}\label{eq:1}
    \begin{aligned}
    {\mathcal{L}_{s}(f_s; x_s, y_s) = 
    -\mathbb{E}_{(x_s,y_s) \in \mathcal{X} \times \mathcal{Y}} \sum_{k=1}^{K} q_k^{ls}log\delta_k(f_s(x_s))}
    \end{aligned}
\end{equation}
where \begin{math}q_k^{ls}\end{math} is obtained by label smoothing \cite{SHOT}.

\noindent
\vspace{0.5em}

\noindent
{\textbf{Domain-Specific Weight Selection: }}
At first, the entire source model is trained with the supervised cross-entropy loss. Once the training completes, we use a procedure of L1-norm-based pruning to prune $1 - p_0$ fraction of the parameters, retaining $p_0$ fraction for the source domain $0$. The weights in each layer of the backbone feature extractor and bottleneck layers are pruned by $1 - p_0$ fraction, while the classifier layer is not. Once pruning is done, we fine-tune the $p_0$ fraction of the model parameters to regain the performance. We refer to these weights as domain-specific weights of the source since they hold the crucial source domain features. After this, the vendor sends the fine-tuned network to the client side for adaptation.

\subsubsection{\textbf{Client-side Adaptation}}\label{III.C.2}
The client-side adaptation is done by optimizing on Information-Maximization (IM) loss \cite{SHOT}. The IM loss is a constitution of $\mathcal{L}_{ent}$ and $\mathcal{L}_{div}$ as follows: 
\begin{equation}
    \begin{aligned}
        \mathcal{L}_{ent}(f_t;X_t) &=    -\mathbb{E}_{x_t\in X_t} \sum\nolimits_{k=1}^{K} \delta_k(f_t(x_t)) \log \delta_k(f_t(x_t)),\\
        \mathcal{L}_{div}(f_t;\mathcal{X}_t) &= \sum\nolimits_{k=1}^{K} \hat{p}_k \log \hat{p}_k \\
        &= D_{KL}(\hat{p}, \frac{1}{K}\mathbf{1}_K) - \log K,
    \end{aligned}
    \label{eq:ent}
\end{equation}
where $f_t(x)=h_t(g_t(x))$ is the $K$-dimensional output of each target sample, $\mathbf{1}_K$ is a $K$-dimensional vector with all ones, and $\hat{p} = \mathbb{E}_{x_t\in\mathcal{X}_t} [\delta(f_t(x_t))]$ is the mean output embedding of the whole target domain. 

\noindent
On the client side, target adaptation is performed keeping the source domain-specific weights frozen. Since domains are encountered sequentially, a client with target domain $t$ might receive a model trained on the previous $t-1$ domains. For each domain, $p_t$ fraction of domain-specific weights is preserved. Hence, the client uses the domain-specific weight selection algorithm to prune weights from the remaining $1 - \sum_{i=0}^{t-1} p_i$ parameters. Once the weights are pruned, fine-tuning is performed to obtain $p_t$ fraction of domain-specific weights for the domain $t$. We also maintain a separate set of batch norm parameters for each domain as prior works \cite{li2016revisiting} indicate that batch norm layers preserve domain-specific characteristics. A parameter mask $M_i$ indicates the set of active weights for a domain with 0s corresponding to the pruned weights and 1s for the active weights. Also, during the training of domain t, the mask $1 - M_{t-1}$ is used to freeze the parameters of previously trained domains.
Hence, in this way, the model keeps training for sequentially arriving client domains,  ensuring that we preserve domain-specific knowledge for each domain to reduce catastrophic forgetting significantly. These masks need to be stored for use during inference, but since these are binary masks, they can be stored efficiently without significant extra memory requirements.

\begin{algorithm}
    \centering
    \small
    \caption{Pruning-aided Continual Domain Adaptation}\label{algo: pacoda}
\begin{algorithmic}[t]
\STATE {\textbf{Input:}\text{ Source model ${f_s = h_s(g_s)}$, Target Data ${\mathcal{D}_{t} = (\mathcal{X}_{t})}$,} {\newline} \text{where ${t \in  }$ \{1, 2,...T\}, n\_epochs\_train, n\_epochs\_finetune}}
\STATE {\textbf{Initialize}\text{${{f_t}}$ to ${{f_s}}$ and freeze ${{h_t}}$}}
\FOR{each domain d}
    \STATE \text{1. Train ${{f_t}}$ for n\_epochs\_train by masking the gradient} \\ \text{\hspace{1.4em}using $1 - M_{t-1}$} \\
    \STATE \text{2. Prune ${1 - {\sum_{i=0}^{t} {{p}_{i}}}}$ fraction of the model using} \\ \text{\hspace{1.4em}L1-pruning and obtain $M_t$}
    \STATE \text{3. Store $M_t$ and Batch Norm layers for current domain $t$}
    \STATE \text{4. Fine-tune ${{f_t}}$ for n\_epochs\_finetune}
    \STATE \text{5. Reset Batch Norm layers to Source Model initialization}
\ENDFOR

\end{algorithmic}
\label{algo: PACoDA}
\end{algorithm}

\noindent
\subsubsection{\textbf{Batch norm based Inference}}
Once the model is trained on a source domain and $\text{T}-1$ target domains, we have $\text{T}$ parameter masks and $\text{T}$ corresponding Batch Norm layers. During inference, a sample from any of the trained domains might be encountered. Hence, the task is to identify the parameter mask and Batch Norm layer to be chosen as the final model to get the best performance on inference samples. For this, we introduce a novel metric \textit{\textbf{B}atch \textbf{N}orm \textbf{S}tatistic \textbf{D}eviation (\textbf{BNSD})}.
Batch Norm layers for each domain contain the running-mean and running-variance statistics. Note that, we only use the mean and variance statistics of each domain, without storing any additional samples for training or inference. BNSD measures the deviation of the inference batch's mean and variance from the stored running statistics associated with the Batch Norm parameters. We propose to use only the model's first layer batch norm since it is closest to the input layer and captures domain-specific information from the input. The BNSD metric is defined as follows:
\begin{multline}\label{bnsd}
        \text{BNSD}({b}_{i}, \text{BN'}, \mathcal{M}'; {f}) = \\
        (\text{Mean}({f}_\mathcal{M'}({b}_{i})) - \text{BN'.running-mean})^2 + \\(\text{Var}({f}_\mathcal{M'}({b}_{i})) - \text{BN'.running-var})^2
\end{multline}
where \begin{math}{{b}_{i}}\end{math} is the \begin{math}{i}^{th}\end{math} batch of inference data, BN' is the first Batch Norm layer with which the BNSD of the input batch is calculated, \begin{math}\mathcal{M'}\end{math} is the  mask corresponding to BN' and \begin{math}{{f}}\end{math} is the final model obtained after training on all domains. \begin{math}{{f}_\mathcal{M'}}\end{math} is the resultant model to be used when the mask \begin{math}\mathcal{M'}\end{math} is applied on the model \begin{math}{{f}}\end{math}. The PaCDA algorithm \ref{algo: pacoda} chooses the Batch Norm layer and corresponding parameter mask which gives the least BNSD for a given batch of samples.

\begin{table*}[ht]
\centering

\resizebox{\textwidth}{!}{
\begin{tabular}{c| cccc | cccc | cccc | cccc}
\hline

       \hline
\hline
  & \multicolumn{4}{c|}{\textbf{A} $\rightarrow$ \textbf{C} $\rightarrow$ \textbf{P} $\rightarrow$ \textbf{R} } & \multicolumn{4}{c|}{\textbf{C} $\rightarrow$ \textbf{A} $\rightarrow$ \textbf{P} $\rightarrow$ \textbf{R}} & \multicolumn{4}{c|}{\textbf{P} $\rightarrow$ \textbf{A} $\rightarrow$ \textbf{C} $\rightarrow$ \textbf{R}}  & 
 
 \multicolumn{4}{c}{\textbf{R} $\rightarrow$ \textbf{A} $\rightarrow$ \textbf{C} $\rightarrow$ \textbf{P}} \\
   \multicolumn{1}{c|}{\textbf{Method}} & \textbf{A} & \textbf{C} & \textbf{P} & \multicolumn{1}{c|}{\textbf{R}} & \textbf{C} & \textbf{A} & \textbf{P} & \multicolumn{1}{c|}{\textbf{R}} & \textbf{P} & \textbf{A} & \textbf{C} & \multicolumn{1}{c|}{\textbf{R}} & \textbf{R} & \textbf{A} & \textbf{C} & \textbf{P} \\
\hline
\hline

\multirow{2}{*}{\textbf{Baseline}} & 61.3 & 54.7 & 68.6 & \multicolumn{1}{c|}{70.8} & 57.7 & 67.7 & 73.9 & \multicolumn{1}{c|}{73.2} & 72.1 & 64.5 & 51.2 & \multicolumn{1}{c|}{70.55} & 74.3 & 65.7 & 55.8 & 72.1\\&\textcolor{Red}{(-12.6)}& \textcolor{Red}{(-1.3)}&
\textcolor{ForestGreen}{(+0.25)} & \textcolor{ForestGreen}{(+0)} & \textcolor{Red}{(-23.2)} & \textcolor{Red}{(-0.6)} & \textcolor{Red}{(-0.4)} & \textcolor{ForestGreen}{(+0)} & \textcolor{Red}{(-20.6)} & \textcolor{Red}{(-3.7)} & \textcolor{Red}{(-1.2)} & \textcolor{ForestGreen}{(+0)} & \textcolor{Red}{(-11.6)} & \textcolor{Red}{(-5.0)} & \textcolor{Red}{(-1.2)} & \textcolor{ForestGreen}{(+0)}\Bstrut \\

\hline

\multirow{2}{*}{\textbf{GSFDA\cite{GSFDA}}} & 72.6 & 55.6 & \textbf{72.7} & \multicolumn{1}{c|}{\textbf{77.2}} & 78.6 & 64.9 & \textbf{72.8} & \multicolumn{1}{c|}{72.4} & 88.6 & 63.1 & 51.5 & \multicolumn{1}{c|}{76.5} & 84.2 & 69.1 & 57.4 & \textbf{80.5}  \Tstrut\\&\textcolor{Red}{(-1.9)}& \textcolor{Red}{(-1.0)}&
 \textcolor{Red}{(-0.3)} & \textcolor{ForestGreen}{(+0)} & \textcolor{Red}{(-3.6)} & \textcolor{Red}{(-0.5)} & \textcolor{Red}{(-0.1)} & \textcolor{ForestGreen}{(+0)} & \textcolor{Red}{(-3.4)} & \textcolor{Red}{(-0.5)} & \textcolor{Red}{(-1.6)} & \textcolor{ForestGreen}{(+0)} & \textcolor{Red}{(-1.8)} & \textcolor{Red}{(-3.3)} & \textcolor{Red}{(-1.7)} & \textcolor{ForestGreen}{(+0)}\Bstrut \\

 \hline

 \multirow{2}{*}{\textbf{Ours(with Domain-id)}} & \underline{73.7} & \underline{56.0} & \underline{71.8} & \multicolumn{1}{c|}{\underline{76.9}} & \underline{80.9} & \underline{67.1} & \underline{72.7} & \multicolumn{1}{c|}{\textbf{76.4}} & \underline{92.9} & \underline{67.2} & \underline{55.8} & \multicolumn{1}{c|}{\textbf{78.1}} & \underline{86.2} & \textbf{73.3} & \underline{58.1} & 79.2 \Tstrut\\&\textcolor{ForestGreen}{(+0)}& \textcolor{ForestGreen}{(+0)}&
 \textcolor{ForestGreen}{(+0)} & \textcolor{ForestGreen}{(+0)} & \textcolor{ForestGreen}{(+0)} & \textcolor{ForestGreen}{(+0)} & \textcolor{ForestGreen}{(+0)} & \textcolor{ForestGreen}{(+0)} & \textcolor{ForestGreen}{(+0)} & \textcolor{ForestGreen}{(+0)} & \textcolor{ForestGreen}{(+0)} & \textcolor{ForestGreen}{(+0)} & \textcolor{ForestGreen}{(+0)} & \textcolor{ForestGreen}{(+0)} & \textcolor{ForestGreen}{(+0)} & \textcolor{ForestGreen}{(+0)} \Bstrut \\

 \hline

 \multirow{2}{*}{\textbf{Ours(w/o Domain-id)}}& \textbf{73.7} & \textbf{56.0} & 71.7 & \multicolumn{1}{c|}{76.1} & \textbf{80.9} & \textbf{67.2} & 72.6 & \multicolumn{1}{c|}{\underline{75.4}} & \textbf{92.9} & \textbf{67.2} & \textbf{55.8} & \multicolumn{1}{c|}{\underline{77.8}} & \textbf{86.2} & \underline{72.8} & \textbf{58.1} & \underline{79.2}\Tstrut\\&\textcolor{ForestGreen}{(+0)}& \textcolor{ForestGreen}{(+0)}&
 \textcolor{Red}{(-0.1)} & \textcolor{ForestGreen}{(+0)} & \textcolor{ForestGreen}{(+0)} & \textcolor{ForestGreen}{(+0.1)} & \textcolor{Red}{(-0.1)} & \textcolor{ForestGreen}{(+0)} & \textcolor{ForestGreen}{(+0)} & \textcolor{ForestGreen}{(+0)} & \textcolor{ForestGreen}{(+0)} & \textcolor{ForestGreen}{(+0)} & \textcolor{ForestGreen}{(+0)} & \textcolor{Red}{(-0.3)} & \textcolor{ForestGreen}{(+0)} & \textcolor{ForestGreen}{(+0)} \Bstrut \\ 

\hline

\end{tabular}
}
\caption{Accuracy (\%) of each method on various Continual Domain Adaptation sequences of Office-Home dataset using ResNet-50 as the backbone. The values in parentheses denote the difference between the model's accuracy at the end of the sequence and the accuracy right after the training of that domain. The lower the value in the parentheses, the higher the catastrophic forgetting. It can be noted that our method performs significantly better at reducing catastrophic forgetting compared to baseline and GSFDA \cite{GSFDA}.}
\label{table:comparison}
\end{table*}

\section{Experiments and Results}
\subsection{Dataset}
\noindent
We demonstrate the efficacy of our approach on the \textbf{Office-Home} dataset \cite{venkateswara2017deep}, which contains a total of 15,500 images belonging to 4 domains with 65 categories each.
We propose to use this dataset since it contains a sizeable inter-domain gap and the effects of catastrophic forgetting would be more evident. 

\subsection{Implementation details}
\noindent
We follow the same experimental setting as SHOT \cite{SHOT} and initialize the backbone with an ImageNet pre-trained ResNet-50 \cite{he2016deep} model. We optimize the loss mentioned in Eq. \ref{eq:1}, Eq. \ref{eq:ent} using Stochastic Gradient Descent with a momentum of 0.9 and weight decay of $1e^{-3}$. We ran our experiments on 12 GB NVIDIA GeForce GTX Titan X GPU. 

\subsection{Comparison with other methods}
\noindent
In Table \ref{table:comparison}, we evaluate our method against other methods in terms of accuracy(\%) at the end of each sequence and catastrophic forgetting relative to the accuracy right after training. The Baseline method (shown in Table \ref{table:comparison}) is a sequential training approach of domains using SHOT\cite{SHOT} method. Since SHOT is well-suited to maximise accuracy on target domains only, 
we observe significant catastrophic forgetting across all sequences. We evaluated both variants of our method - with Domain-id (having access to domain-ID of inference samples) and w/o domain-ID (using BNSD metric to map the inference batch to the appropriate domain-specific parameter mask). We also compare our results with GSFDA\cite{GSFDA}, which reported Continual DA results for the same four Office-Home sequences. 
We also use the same 0.8/0.2 train-test split on the source domain as GSFDA, and source-domain accuracies are reported on the test split.
Our method not only achieves higher accuracy than GSFDA, but even without using domain-id, PaCDA significantly reduces catastrophic forgetting. 

\noindent
Our inference process, using the BNSD metric, selects the suitable mask and batch norm layer almost perfectly, and the performance of our PaCDA method without domain-id is almost as good as PaCDA with domain-id method.

\noindent
In Table \ref{table:domain_params_ablation}, for the same set of sequences, we evaluate the accuracy of each domain when using different parameter masks (and their corresponding batch norm layers) on the final model obtained at the end of the sequence. As expected, the corresponding parameter mask for each domain gives the best results. One intriguing result to note is, in Table \ref{table:domain_params_ablation} sequence CAPR, the accuracy on domain A using parameter mask of R is slightly higher than the accuracy obtained using parameter mask of A. This also reflects in Table \ref{table:comparison} sequence CAPR, when inferred on domain A, PaCDA w/o Domain-id accuracy is higher than with domain-id. 


\begin{table}[!t]
\centering

 \vskip 0.15in
\begin{adjustbox}{width=\columnwidth,center}
\begin{tabular}{c | c  c  c  c || c | c c c c } 
\hline
 & \textbf{A} & \textbf{C} & \textbf{P} & \textbf{R} & & \textbf{C} & \textbf{A} & \textbf{P} & \textbf{R}\\
\hline
{\textbf{A}} & \underline{\textbf{73.7}} & 44.1 & 64.5 & 73.1 & \textbf{C} & \underline{\textbf{80.9}} & 50.8 & 60.1 & 63.7\\
{\textbf{C}} & 67.1 & \underline{\textbf{56.0}} & 60.4 & 67.7 & \textbf{A} & 70.33 & \textbf{67.1} & 65.5 & 69.5 \\
{\textbf{P}} & 66.3 & 53.0 & \underline{\textbf{71.8}} & 73.3 & \textbf{P} & 72.3 & 66.7 & \underline{\textbf{72.7}} & 75.5\\
{\textbf{R}} & 68.1 & 53.1 & 71.4 & \underline{\textbf{76.9}} & \textbf{R} & 71.0 & \underline{67.6} & 71.7 & \underline{\textbf{76.4}}\\
\hline \hline
& \textbf{P} & \textbf{A} & \textbf{C} & \textbf{R} & & \textbf{R} & \textbf{A} & \textbf{C} & \textbf{P}\\
\hline
{\textbf{P}} & \underline{\textbf{92.9}} & 52.1 & 41.7 & 73.0 & \textbf{R} & \underline{\textbf{86.2}} & 64.8 & 45.8 & 76.7 \\
{\textbf{A}} & 84.5 & \underline{\textbf{67.2}} & 44.1 & 74.2 & \textbf{A} & 82.6 & \underline{\textbf{73.3}} & 48.7 & 75.1 \\
{\textbf{C}} & 80.0 & 63.0 & \underline{\textbf{55.8}} & 69.9 & \textbf{C} & 78.0 & 67.4 & \underline{\textbf{58.1}} & 70.2 \\
{\textbf{R}} & 84.3 & 66.6 & 52.8 & \underline{\textbf{78.1}} & \textbf{P} & 80.5 & 68.2 & 55.2 & \underline{\textbf{79.2}} \\

\end{tabular}
 \end{adjustbox}
 
 \captionsetup{width=1.0\linewidth}
 \caption{Accuracy of each domain data when inferred using different domain-specific weights for four sequences on the Office-Home dataset. Each quadrant is a sequence in which parameter mask changes along the rows and input domain changes along the column. In each column of a sequence, the bold value is the expected best accuracy, and the underlined value is the actual best accuracy obtained.}
 \label{table:domain_params_ablation}
\vspace{-1.5em}
 \end{table}

\section{Conclusion}
\noindent
In this work, we address the practical problem of Continual Domain Adaptation using a pruning-aided approach (PaCDA). At first, we investigate and find that pruning model parameters provides room for storing crucial domain-specific knowledge. We introduce a novel domain-specific weight selection algorithm for multiple client domains and reduce catastrophic forgetting of past-seen domains to a vast extent. We use a novel batch norm based metric to infer on test domain batches during inference. Hence, we not only achieve state-of-the-art results on standard DA benchmarks, but our results also demonstrate that appropriate model parameters are chosen during inference. As a part of future analysis, we plan to extend our work using model expansion approaches to enable the learning of a larger number of domains sequentially.

\section{Acknowledgement}
\noindent
This work was supported by WIRIN (Wipro IISc Research and Innovation Network) project.

\FloatBarrier

{\small
\bibliographystyle{ieee_fullname}
\bibliography{egbib}
}

\end{document}